\documentclass[acmsmall]{acmart}

\usepackage{tcolorbox}
\usepackage{float}
\usepackage{hyperref}
\usepackage{geometry}
\usepackage{multirow,bigstrut}
\usepackage{colortbl}
\usepackage{longtable}
\usepackage{changepage}
\usepackage[printonlyused]{acronym} 
\usepackage{parskip}
\usepackage{ulem}
\usepackage{pdfpages}
\usepackage{enumitem}
\usepackage{booktabs, multirow} 
\usepackage{soul}
\usepackage{lineno} 

\definecolor{Gray1}{gray}{0.95}
\definecolor{Gray2}{gray}{0.89}

\AtBeginDocument{%
  \providecommand\BibTeX{{%
    \normalfont B\kern-0.5em{\scshape i\kern-0.25em b}\kern-0.8em\TeX}}}

\begin{document}

\title[]{Labeling instructions matter in biomedical image analysis}


\author{Tim Rädsch†}
\affiliation{%
  \institution{German Cancer Research Center (DKFZ)}
  \department{Div. Intelligent Medical Systems and HI Helmholtz Imaging}
  \city{Heidelberg}
  \country{Germany}
}

\author{Annika Reinke}
\affiliation{%
  \institution{German Cancer Research Center (DKFZ)}
  \department{Div. Intelligent Medical Systems and HI Helmholtz Imaging}
  \city{Heidelberg}
  \country{Germany}
}
\affiliation{%
  \institution{Heidelberg University}
  \department{Faculty of Mathematics and Computer Science}
  \city{Heidelberg}
  \country{Germany}
}

\author{Vivienn Weru}
\affiliation{%
  \institution{German Cancer Research Center (DKFZ)}
  \department{Div. Biostatistics}
  \city{Heidelberg}
  \country{Germany}
}

\author{Minu D. Tizabi}
\affiliation{%
  \institution{German Cancer Research Center (DKFZ)}
  \department{Div. Intelligent Medical Systems and HI Helmholtz Imaging}
  \city{Heidelberg}
  \country{Germany}
}

\author{Nicholas Schreck}
\affiliation{%
  \institution{German Cancer Research Center (DKFZ)}
  \department{Div. Biostatistics}
  \city{Heidelberg}
  \country{Germany}
}

\author{A. Emre Kavur}
\affiliation{%
  \institution{German Cancer Research Center (DKFZ)}
  \department{HI Applied Computer Vision Lab, Division of Medical Image Computing; Div. Intelligent Medical Systems}
  \city{Heidelberg}
  \country{Germany}
}

\author{Bünyamin Pekdemir}
\affiliation{%
  \institution{Helmholtz Zentrum München}
  \department{Helmholtz Pioneer Campus}
  \city{München}
  \country{Germany}
}

\author{Tobias Roß}
\affiliation{%
  \institution{German Cancer Research Center (DKFZ)}
  \department{Div. Intelligent Medical Systems}
  \city{Heidelberg}
  \country{Germany}
}
\affiliation{%
  \institution{Quality Match GmbH}
  \department{}
  \city{Heidelberg}
  \country{Germany}
}

\author{Annette Kopp-Schneider*}
\affiliation{%
  \institution{German Cancer Research Center (DKFZ)}
  \department{Div. Biostatistics}
  \city{Heidelberg}
  \country{Germany}
}

\author{Lena Maier-Hein}
\authornote{Shared last authors.}
\authornote{e-mail: \{tim.raedsch, l.maier-hein\}@dkfz-heidelberg.de.}
\email{l.maier-hein@dkfz.de}
\affiliation{%
  \institution{German Cancer Research Center (DKFZ)}
  \department{Div. Intelligent Medical Systems and HI Helmholtz Imaging}
  \city{Heidelberg}
  \country{Germany}
}
\affiliation{%
  \institution{Heidelberg University}
  \department{Faculty of Mathematics and Computer Science and Medical Faculty}
  \city{Heidelberg}
  \country{Germany}
}
\affiliation{%
  \institution{National Center for Tumor Diseases (NCT)}
  \city{Heidelberg}
  \country{Germany}
}

\renewcommand{\shortauthors}{Tim Rädsch et al.}

\begin{abstract}
\textbf{Abstract:} 
Biomedical image analysis algorithm validation depends on high-quality annotation of reference datasets, for which labeling instructions are key. Despite their importance, their optimization remains largely unexplored. Here, we present the first systematic study of labeling instructions and their impact on annotation quality in the field. Through comprehensive examination of professional practice and international competitions registered at the MICCAI Society, we uncovered a discrepancy between annotators’ needs for labeling instructions and their current quality and availability. Based on an analysis of 14,040 images annotated by 156 annotators from four professional companies and 708 Amazon Mechanical Turk (MTurk) crowdworkers using instructions with different information density levels, we further found that including exemplary images significantly boosts annotation performance compared to text-only descriptions, while solely extending text descriptions does not. Finally, professional annotators constantly outperform MTurk crowdworkers. Our study raises awareness for the need of quality standards in biomedical image analysis labeling instructions.

\end{abstract}


\maketitle
\setlength{\parskip}{0.5em}
\newpage

\section*{Introduction}
\label{sec:introduction}
\ac{ML} is in the process of revolutionizing medicine, with \ac{DL} as a key enabling technology~\cite{benjamens_state_2020}. High-quality annotated datasets are a critical bottleneck for supervised \ac{DL} and the quality of the annotated data is crucial for algorithm performance~\cite{shad_designing_2021,peiffer-smadja_machine_2020,hu_challenges_2020,willemink_preparing_2020}. Recent work reflects an increasing awareness of widespread problems in commonly used image benchmarks, which are subject to errors~\cite{northcutt_pervasive_2021,radsch_what_2021} and biases~\cite{paullada_data_2021,peng_mitigating_2021}, and calls for a fundamental change in dataset culture~\cite{noauthor_rise_2022}. Annotation-related problems may be particularly relevant in the field of biomedical image analysis, where data is typically sparse~\cite{maier-hein_surgical_2022}, inter-rater variability is naturally high~\cite{joskowicz_inter-observer_2019,freeman_iterative_2021}, labeling ambiguities occur~\cite{kohli_medical_2017} and medical experts have their individual style of annotations~\cite{balagopal_psa-net_2021,freeman_iterative_2021}.

In addition to these limitations, domain expert resources are typically limited and costly~\cite{freeman_iterative_2021}. As a result, an increasingly popular approach to generating image annotations involves outsourcing the labeling task to crowdsourcing platforms~\cite{orting_survey_2020,crequit_mapping_2018} or professional labeling companies~\cite{maier-hein_surgical_2022}. Historically, outsourcing was first performed on general labor markets like \ac{MTurk}~\cite{noauthor_amazon_nodate}, which is still the predominant choice in health research ~\cite{crequit_mapping_2018}. With rising demand for annotations, professional labeling companies catering to the specific needs of their target domain emerged. An overview of data annotation from a surgical data science perspective is provided by Maier-Hein et al.~\cite{maier-hein_surgical_2022}.

While crowdsourcing has successfully been applied in a number of medical imaging applications, the high variation in labeling quality is a key issue~\cite{budd_can_2021,orting_survey_2020,heim_large-scale_2018,cheplygina_early_2016,maier-hein_can_2014}. Poor annotation quality can generally be attributed to two main factors:

\begin{enumerate}
    \item Lack of motivation and time pressure: Driven by subpar compensation policies~\cite{litman_relationship_2015} and power dynamics~\cite{denton_whose_2021}, \ac{MTurk} suffers from workers that perform sloppy annotations with the goal of completing tasks as fast as possible and thus maximizing the monetary reward. This has led to a notable decline in the annotation quality in recent years~\cite{kennedy_shape_2020}.
    \item Lack of knowledge/expertise: A worker depends on the provided information in order to create the desired outcome for a given task~\cite{hossfeld_crowdsourcing_2014,tokarchuk_analyzing_2012}. A lack of labeling instructions leads to workers filling knowledge gaps with their own interpretations~\cite{clark_grounding_1991}.
\end{enumerate}

While the first problem has been addressed in literature~\cite{sullivan_deep_2018,heim_large-scale_2018,albarqouni_playsourcing_2016,mavandadi_distributed_2012,luengo-oroz_crowdsourcing_2012}, the notion of training workers has been given almost no attention in the context of medical imaging~\cite{orting_survey_2020,crequit_mapping_2018}. Regardless of the annotator type (general crowdsourcing or professional labeling companies), knowledge transfer is typically achieved via so-called labeling instructions, which specify how to correctly label the (imaging) data. Such labeling instructions are not only needed for training non-experts but can also help reduce experts’ inter-rater variability by defining a shared standard.

Given the importance of annotation quality for algorithm performance and the fundamental role of labeling instructions in this process, it may be surprising that extremely limited effort has been invested into the question of how to best generate labeling instructions in a way that annotation quality is maximized. While previous work on instructions has dealt with instructions for annotation systems with a focus on Natural Language Processing tasks~\cite{ning_easy_2020,k_chaithanya_manam_taskmate_2019,bragg_sprout_2018,manam_wingit_2018,chang_revolt_2017} and standardization of dataset reporting~\cite{gebru_datasheets_2021,maier-hein_bias_2020}, we are not aware of any work dedicated to labeling instructions in the field of biomedical image analysis and involving professional annotation companies. Closing this gap in the literature, our contribution is two-fold:

\begin{figure}[H]
    \centering
    \includegraphics[width=0.95\linewidth]{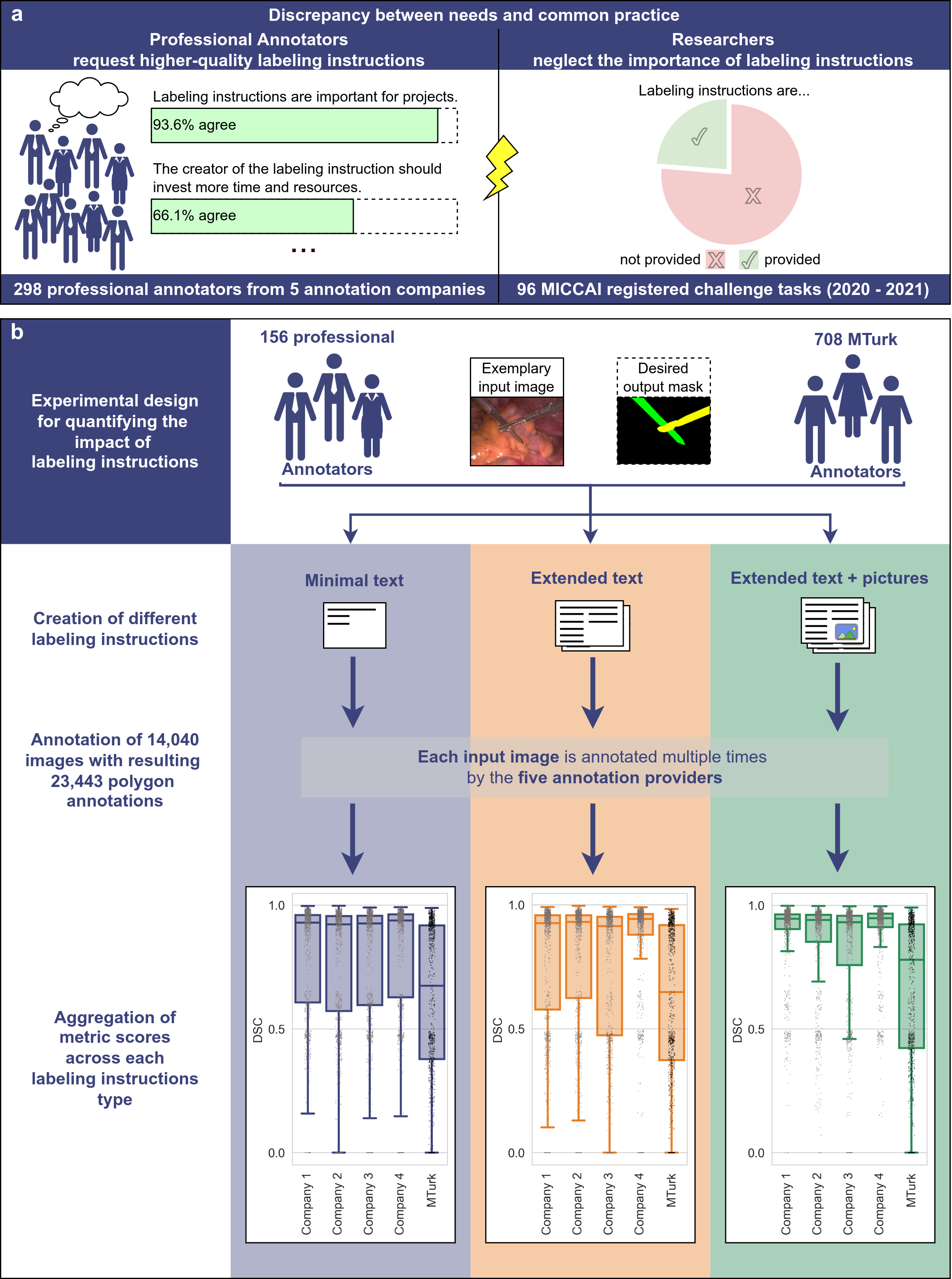}
    \caption{\textbf{Overview of the methodology.} \textbf{a} The field of biomedical image analysis suffers from a notable discrepancy between the needs of those annotating the data and the actual availability of labeling instructions (if any). \textbf{b} Experimental design for quantifying the impact of labeling instructions. Three types of labeling instructions were created for the same dataset, namely: a) \textit{minimal text} , b) \textit{extended text} and c) \textit{extended text including pictures}. A total of 864 annotators, recruited from the popular crowdsourcing company \acl{MTurk} (\acs{MTurk}) and from four professional labeling companies then annotated a dataset of 14,040 surgical images based on one of the three description types. A two-part beta mixed model was used for quantifying the impact of the labeling instructions and the other covariates.}
    \label{fig:experiment_overview}
\end{figure}

\begin{enumerate}
    \item Analysis of common practice (Fig.~\ref{fig:experiment_overview}a): Through comprehensive examination of professional annotating practice and major international biomedical image analysis competitions, we systematically investigated the importance of labeling instructions, their quality and their availability.
    \item Experiments on the impact of biomedical image labeling instructions (Fig.~\ref{fig:experiment_overview}b): Based on 14,040 images annotated by a total of 864 annotators from five different annotation providers, we experimentally determined the effect of varying information density in labeling instructions. In this context, we also investigated annotation quality of professional annotators in comparison to the current status quo in scalable annotations, \ac{MTurk} crowdworkers. 

\end{enumerate}

\section*{Results}
\label{sec:results}
Given the lack of (1) awareness of the importance of labeling instructions and (2) quantitative research investigating how to best perform the labeling, we initiated our study by systematically analyzing the perspective and work characteristics of professional annotators, and common practice of labeling instructions in leading biomedical imaging competitions. Subsequently, we investigated the impact of labeling instructions with varying levels of information density on the annotation quality, and the effect of different annotator types on biomedical imaging data.

\subsection*{Professional annotators request better labeling instructions}
\begin{figure}[H]
    \centering
    \includegraphics[width=\linewidth]{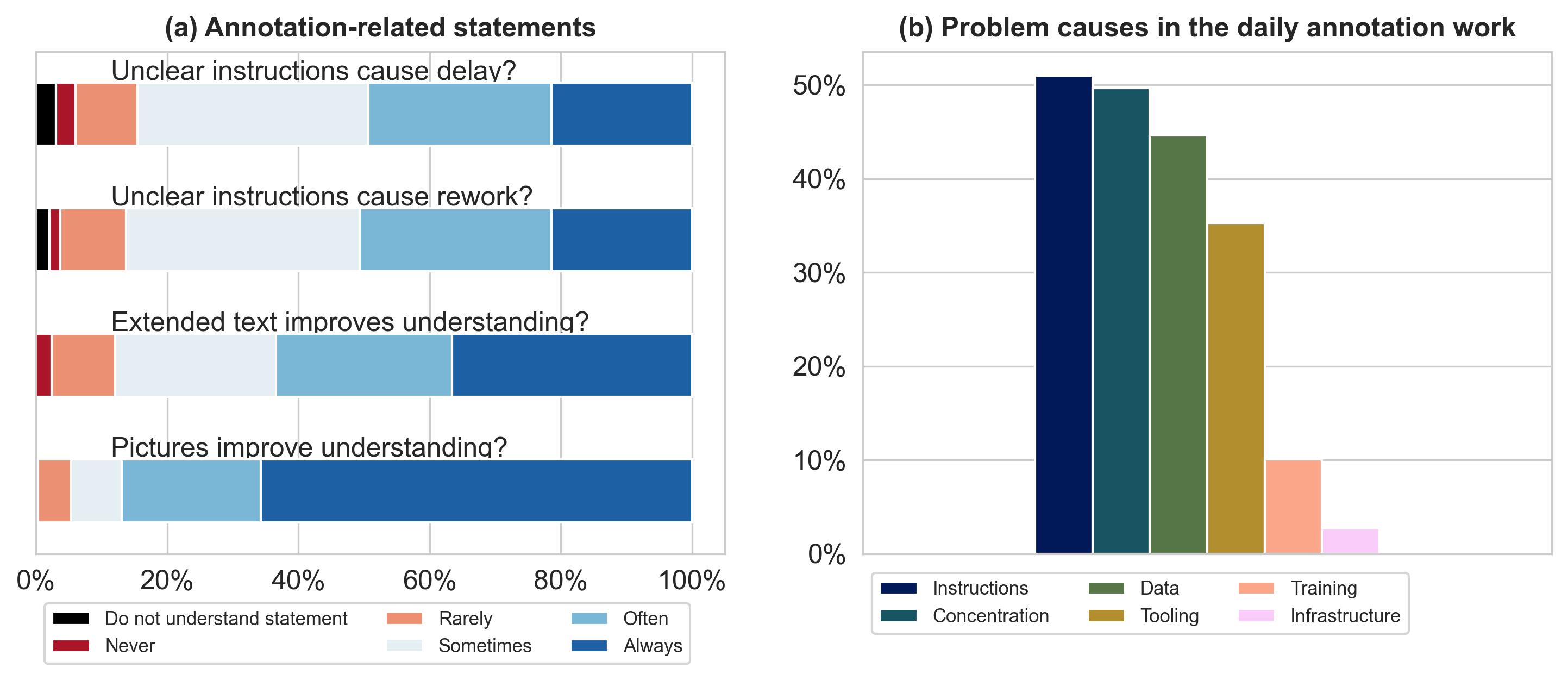}
    \caption{\textbf{Labeling instructions, in their current state, are a major driver for problems, delay and rework during the labeling of data.} \textbf{a} Professional annotators agree that unclear instructions consistently cause delay and rework. Comprehensive text descriptions and images are perceived to improve annotation quality. \textbf{b} Professional annotators attribute labeling instructions as the primary cause for problems related to their daily annotation work, followed by concentration issues and poor input data. Answers were processed from 298 annotators from five annotation companies.}
    \label{fig:overview_survey_prof_ann}
\end{figure}

To motivate our empirical study on annotation quality, we conducted an international survey among 363 (298 after filtering noisy answers) professional annotators employed by five different internationally operating annotation companies. Depicted in Fig.~\ref{fig:overview_survey_prof_ann}, the results  reveal that the majority of annotators request more time and resources to be spent in the generation of labeling instructions. In fact, poor labeling instructions were identified as the primary cause of problems related to annotation work followed by concentration issues (50\%) and poor input data (45\%).

\subsection*{The importance of labeling instructions may be underrated}
Despite their apparent importance, earlier research revealed that labeling instructions are typically not provided and/or reported in the field of biomedical image analysis~\cite{maier-hein_why_2018}. This even holds true for international image analysis competitions although these can be expected to provide particularly high quality with respect to validation standards. To address this issue, the \ac{MICCAI}, the largest international society in the field, took action and developed a comprehensive reporting guideline~\cite{maier-hein_bias_2020} for biomedical image analysis competitions. The guideline comprises an entire paragraph on reporting the annotation process, including the labeling instructions. Prior to conducting a competition in the scope of a \ac{MICCAI} conference, researchers must put the report for their competition online~\cite{medical_image_computing_and_computer_assisted_intervention_call_nodate} to foster transparency and reproducibility, and to prevent cheating~\cite{reinke_how_2018}. To capture the state of the art regarding labeling instructions in biomedical image analysis, we analyzed all \ac{MICCAI} competitions officially registered in the past two years (PRISMA statement~\cite{page_prisma_2021}: Supp.~\ref{supp:prisma_statement}). Although the reporting guideline explicitly asks for (a link to) the labeling instructions, 76\% of the recent \ac{MICCAI} competitions do not report any labeling instructions (cf. Fig.~\ref{fig:experiment_overview}a, right side). Given that \ac{MICCAI} competitions make up around 50\% of the biomedical image analysis competitions in a year~\cite{maier-hein_why_2018}, this can be regarded as a widely spread phenomenon.

In current biomedical image analysis practice, labeling instructions are thus often neither of sufficient quality, nor are they appropriately reported and valued in the scientific community. Both issues negatively impact scientific quality in the field.

\subsection*{Comprehensive text descriptions do not boost annotation performance}

\begin{figure}[]
    \centering
    \includegraphics[width=\linewidth]{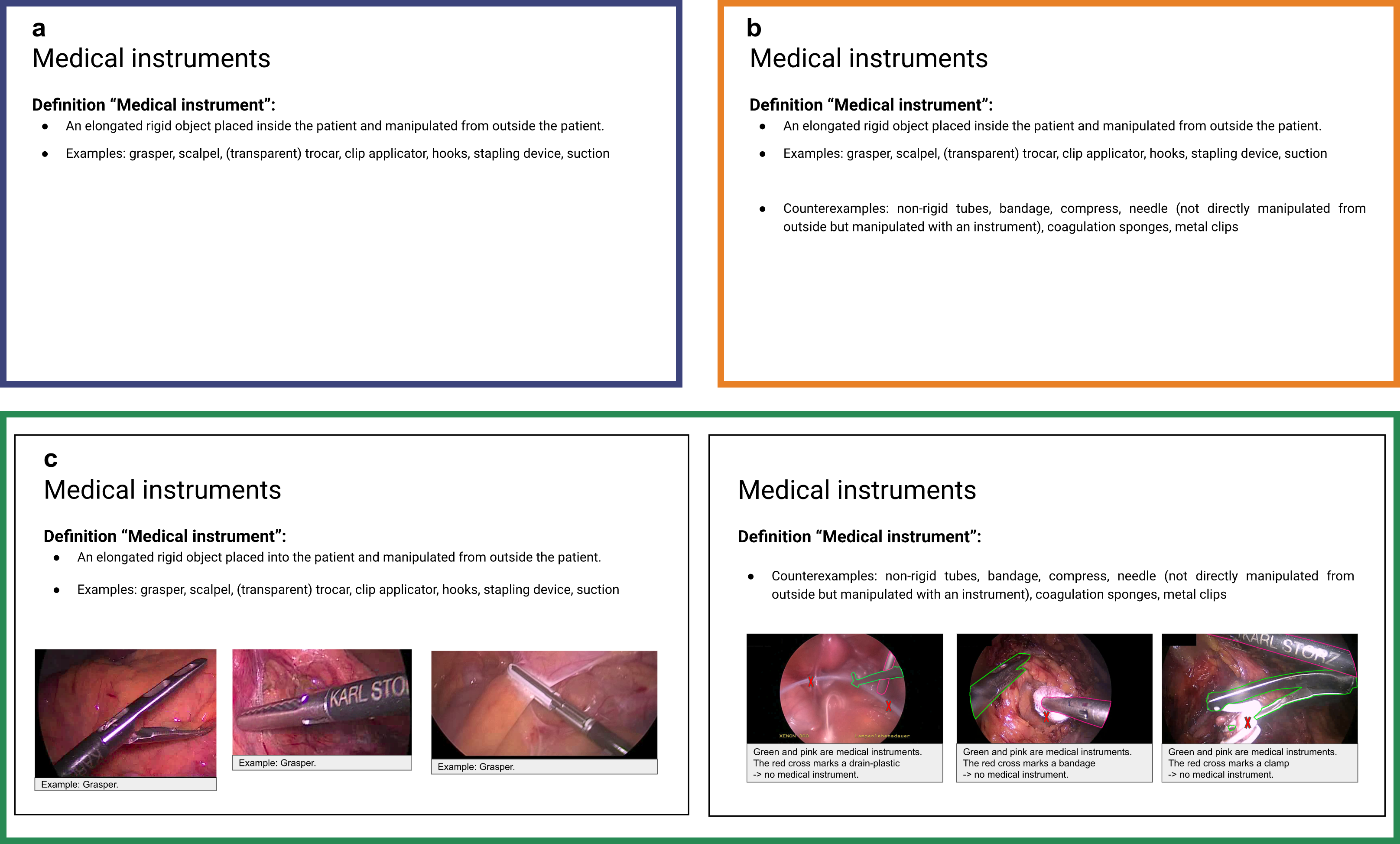}
    \caption{\textbf{Example of labeling instructions.} Medical instruments are initially defined in the \textit{minimal text labeling} instructions (\textbf{a}). The definition is deepened in the \textit{extended text labeling }instructions (\textbf{b}) and enriched with images in the \textit{extended text including pictures labeling} instructions (\textbf{c}).}
    \label{fig:example_slides_information_labeling_instructions}
\end{figure}

The shortcomings in quality we found in common practice regarding labeling instructions called for an investigation on how this quality can be improved. As a first step in this direction, we sought to determine the impact of different types of labeling instructions on the quality of annotation of a particular dataset. The selected dataset~\cite{maier-hein_heidelberg_2021,ros_comparative_2021}, which can be handled by crowdworkers, combines highest quality reference annotations and 21 meta annotations per annotated image that reflect the annotation difficulty. We created three distinct types of labeling instructions with varying levels of information density, namely (a) \textit{minimal text}, (b) \textit{extended text}, and (c) \textit{extended text including pictures}, as detailed in \nameref{sec:method} and Supp.~\ref{supp:li_type_one}-\ref{supp:li_type_three}. Example instructions are provided in Fig.~\ref{fig:example_slides_information_labeling_instructions}. The obtained 23,443 annotations on 14,040 images from the 864 annotators were analyzed with a two-part \ac{ZIBMM}.

In contrast to the \textit{limited text labeling} instructions, the \textit{extended text labeling} instructions included more detailed descriptions, counterexamples and information on uncommon annotation cases that might appear. We define an annotation with a metric score equal to zero as a severe annotation error. For the 4,680 images annotated with the \textit{extended text labeling} instructions, we observed a minor increase in the number of severe annotation errors compared to the \textit{limited text labeling} instructions [median: +0.4\%; max: +14.8\%; min: -31.7\%]. Furthermore, we observed no impact on the median \ac{DSC}~\cite{dice_measures_1945}, and only a minor increase in the \ac{IQR} [median: +1.8\%; max: +33.3\%; min: -75.8\%]. These results contradict the initial assessment of the professional annotators. The absent effect of the \textit{extended text labeling} instructions is reinforced by the results of the two-part \ac{ZIBMM}, where we obtained no statistically significant difference for the \textit{extended text labeling} instructions compared to minimal text from both the first and second part of the model, revealing that extended text descriptions do not boost annotation performance.

\subsection*{Exemplary images are crucial for high quality annotations}
\begin{figure}[]
    \centering
    \includegraphics[width=\linewidth]{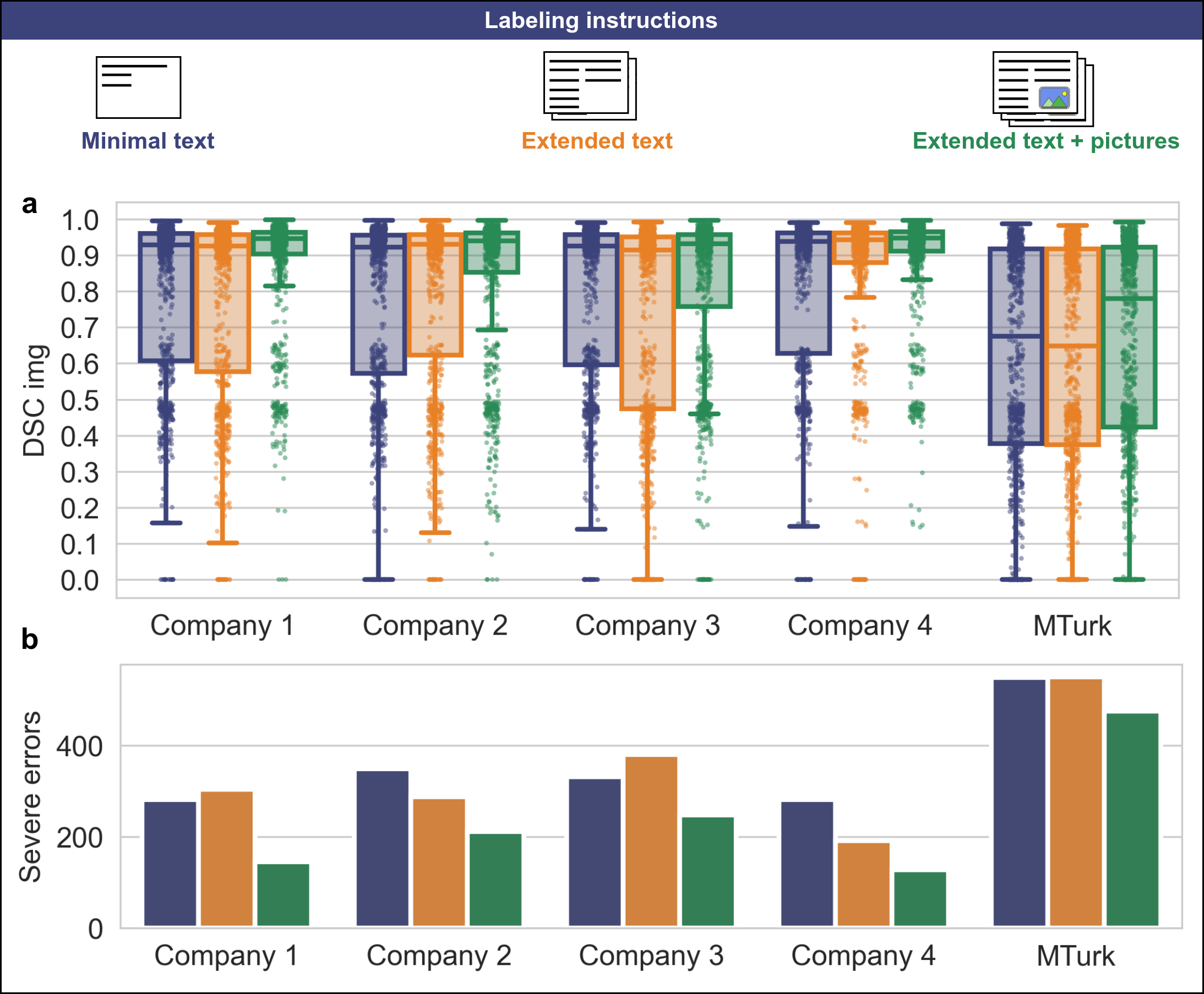}
    \caption{\textbf{Key findings of our study:} (1) Extended text descriptions (orange) do not necessarily boost annotation performance compared to minimal text descriptions (blue), while (2) including images (green) gives a clear benefit for all annotation providers. (3) Professional labeling companies (Companies 1 - 4) provide substantially higher-quality annotations compared to the most popular crowdsourcing platform \acl{MTurk} (\acs{MTurk}). The \acl{DSC} (\acs{DSC}) score has been aggregated for each annotated image and is displayed aggregated for each pair of company and labeling instruction as a dots- and boxplot (the band indicates the median, the box indicates the first and third quartiles and the whiskers indicate $ \pm 1.5 \times$ interquartile range). The absolute number of severe annotation errors, defined as annotations with a metric score equal to zero, is also shown. A total of 14,040 images were annotated by 156 annotators from four professional companies and 708 \ac{MTurk} crowdworkers.}
    \label{fig:results_overview_dsc_and_invalid_annotations}
\end{figure}

Professional annotators claim that pictures help them understand labeling instructions (cf. Fig.~\ref{fig:overview_survey_prof_ann}a). Therefore, the \textit{extended text including pictures labeling} instructions were enriched by pictures including rare occurrences. In comparison to the \textit{extended text labeling} instructions, the number of severe annotation errors was reduced for all five annotation providers [median: -33.9\%; max: -13.6\%; min: -52.3\%]. Furthermore, their median \ac{DSC} score increased [median: +2.2\%; max: 20.0\%; min: +1.1\%], and their IQR was reduced [median: -58.3\% ; max: -9.1\%; min: -84.2\%] (cf. Fig.~\ref{fig:results_overview_dsc_and_invalid_annotations}a). This reinforces professional annotators’ initial assessment that pictures improve their understanding (cf. Fig.~\ref{fig:overview_survey_prof_ann}a). The improvements occurred mainly on the difficult annotation cases (cf. Supp.~\ref{supp:dsc_simple_and_chaos_category}). Based on the two-part \ac{ZIBMM}, the odds of obtaining a severe annotation error with these labeling instructions are 0.37 times (\ac{CI}: 0.28, 0.50) that with \textit{minimal text labeling} instructions. From the second part of the two-part \ac{ZIBMM}, we obtained no significant difference in the \ac{DSC} score, once an object was identified. Thus, the improvements primarily stemmed from the additional reduction of severe annotation errors  (cf. Fig.~\ref{fig:results_overview_dsc_and_invalid_annotations}b).

\subsection*{Professional labeling companies outperform the most popular crowdsourcing platform}

\begin{figure}[]
    \centering
    \includegraphics[width=\linewidth]{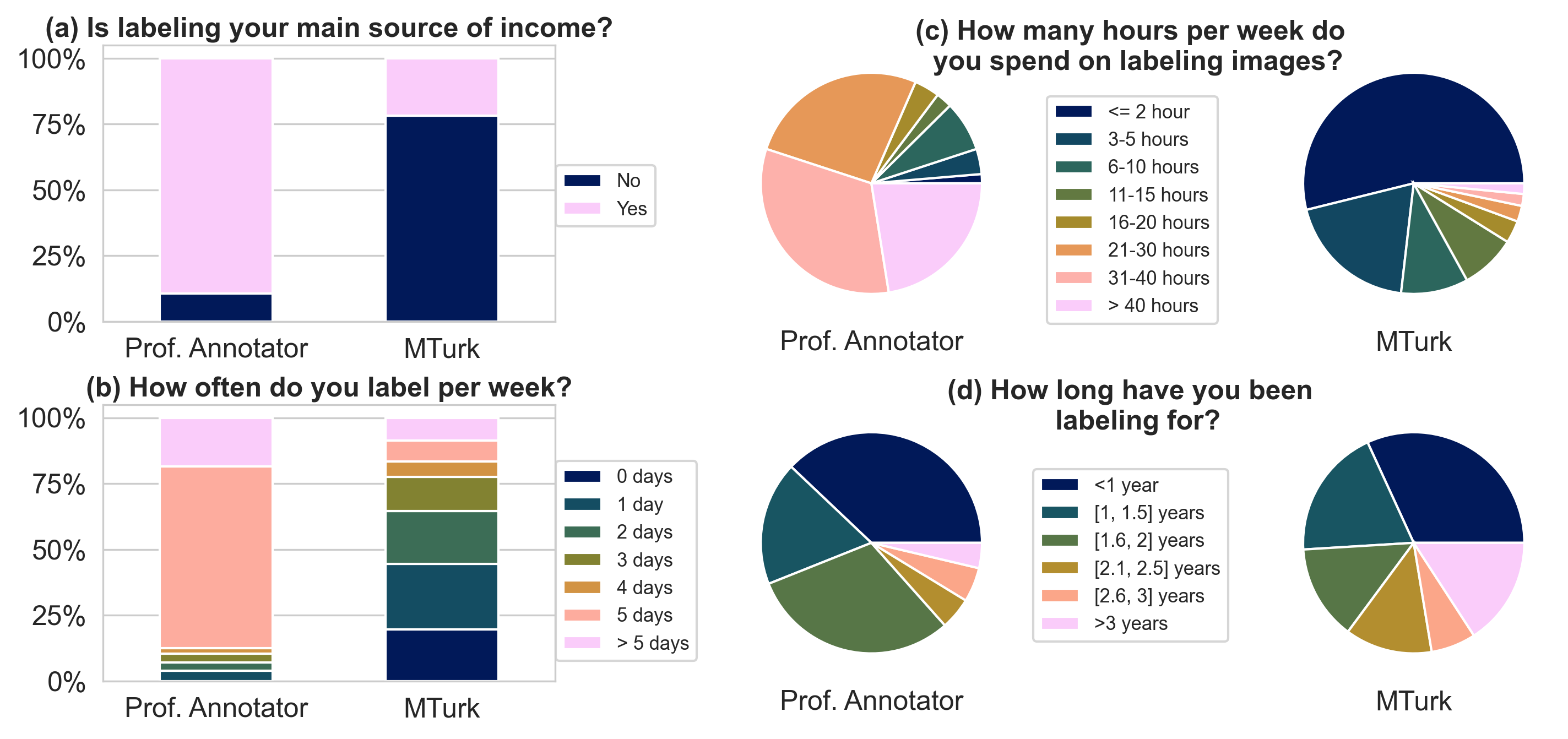}
    \caption{\textbf{Work characteristics of professional annotators and \acl{MTurk} (\acs{MTurk}) crowdworkers.} In comparison to the \ac{MTurk} crowdworkers, professional annotators conduct labeling as their main source of income (\textbf{a}), label more often in a week (\textbf{b}), and spend a higher number of hours per week on labeling images (\textbf{c}). In contrast, \ac{MTurk} crowdworkers have a longer employment history with labeling than professional annotators (\textbf{d}). Answers were processed from 298 professional annotators from five annotation companies and 518 \ac{MTurk} crowdworkers. }
    \label{fig:work_characteristics}
\end{figure}

In comparison to \ac{MTurk} crowdworkers, professional annotators conduct labeling as their main source of income, label more often in a week, and label for a higher number of weekly hours (cf. Fig.~\ref{fig:work_characteristics}a - c). In contrast, \ac{MTurk} workers have a longer employment history in labeling than professional annotators, as displayed in detail in Fig.~\ref{fig:work_characteristics}d. This observation is consistent with the historical development of the data annotation market, where general labor markets, including \ac{MTurk}, preceded professional companies. Given \textit{minimal text labeling} instructions, professional annotators produced less severe annotation errors [companies median: 307; \ac{MTurk} median: 549]. Furthermore, the professional annotators’ annotations generated a higher median \ac{DSC} score [companies median: 0.93; \ac{MTurk} median: 0.67], and a smaller \ac{IQR} of the \ac{DSC} score [companies median: 0.36; \ac{MTurk} median: 0.67]. Both annotator types only showed minor or no improvement with the \textit{extended text labeling} instructions. While both annotator types benefitted from added pictures, professional annotators displayed a stronger reduction of severe annotation errors [companies median: -34.4\%;  \ac{MTurk} median; -13.6\%] and of the \ac{IQR} [companies median: -63.00\%; \ac{MTurk} median: -9.1\%]. In contrast, \ac{MTurk} crowdworkers displayed a stronger improvement of the median \ac{DSC} score [companies median: +1.7\%; \ac{MTurk} median: +20.0\%]. Under the same conditions, the odds of severe annotation errors for a professional annotator were 0.09 times (\ac{CI}: 0.06, 0.12) that of a \ac{MTurk} crowdworker, keeping all other factors constant. Similarly, once an object was identified, the odds for a professional annotator of achieving a perfect \ac{DSC} score were 94.7\% (1.947, \ac{CI}: 1.69, 2.24) higher than those of a \ac{MTurk} crowdworker.

\section*{Discussion}
\label{sec:discussion}
To our knowledge, this study is the first to quantitatively and critically examine the role of labeling instructions in professional and crowdsourced annotation work. We were able to uncover a major discrepancy between their importance and quality/availability, and determine which type of labeling instructions is the most effective. While labeling instructions play a crucial role in the creation of biomedical image analysis datasets, current common practice is insufficient and neither meets the requirements of industry (e.g. in creating large-scale datasets), nor those of academia (e.g. in hosting competitions) (cf. Fig.~\ref{fig:experiment_overview}a). Notably, professional annotators demand a higher time and resource commitment from the creators of labeling instructions. Furthermore, they identify current labeling instructions as a main cause for annotation delay and rework (cf. Fig.~\ref{fig:overview_survey_prof_ann}a). Interestingly, we found that extended text descriptions do not necessarily boost annotation performance compared to minimal text descriptions (cf. Fig.~\ref{fig:results_overview_dsc_and_invalid_annotations}), although professional annotators expect them to improve the understanding of the labeling tasks (cf. Fig.~\ref{fig:overview_survey_prof_ann}a). In contrast, the addition of pictures resulted in a clear improvement among all annotation providers (cf. Fig.~\ref{fig:results_overview_dsc_and_invalid_annotations}), which matches the assessment of the professional annotators (cf. Fig.~\ref{fig:overview_survey_prof_ann}a). This improvement was mainly observed on ambiguous images with challenging conditions, such as poor illumination or intersecting objects, as depicted in Supp.~\ref{supp:dsc_simple_and_chaos_category}. Since \ac{ML} models fail their prediction more often on ambiguous images than on clear images~\cite{shad_designing_2021}, the correct annotation of such images in training datasets is particularly crucial for good performance. Lastly, annotators from professional companies provide substantially higher-quality annotations compared to those from the most popular crowdsourcing platform in health research, \ac{MTurk}, regardless of the type of labeling instruction.

One of the key implications of our findings is that there is a huge discrepancy between the potential impact of labeling instructions on generating the desired annotations and their role in current (research) practice. Our study shows that scope and specific design choices of labeling instructions play a major role in generating the desired annotations. Among other contributing factors such as prior annotation expertise or training, labeling instructions may thus determine the attention to detail that annotators pay to annotating challenging images. The importance of this is amplified in cases of datasets becoming increasingly complex and diversified over time, where more and more challenging data points are added due to initial prediction difficulties of the ML model. Consequently, researchers and practitioners alike should ensure proper representation of the necessary information in their labeling instructions, extend them if needed, and invest the necessary time to produce informative annotated images. However, we observed that 76\% of the recent \ac{MICCAI} competitions did not report their labeling instructions (Fig.~\ref{fig:experiment_overview}a). Since competitions are aimed at generating high publicity, it can be assumed that their current handling of labeling instructions represents the upper bound of quality regarding common practice, with the quality being significantly lower for research projects devoid of intense public scrutiny. This discrepancy is alarming and calls for a paradigm shift in common practice.

Another implication of our study is that generating and publishing labeling instructions is a precondition for enabling independent verification and reproduction of the created annotations. Similarly to how published code enables the verification of algorithmic results in research papers, access to labeling instructions is necessary to understand annotators’ decisions and potentially recreate annotations. Furthermore, the information provided in labeling instructions is important for \ac{ML} practitioners. Based on the annotation decisions (e.g. handling of occluded objects of interest) implemented in a dataset, \ac{ML} practitioners need to define their own desired outcome for these occurrences and modify their \ac{ML} model accordingly. Given their impact on the resulting annotations and the current poor state of datasets~\cite{noauthor_rise_2022,northcutt_pervasive_2021,radsch_what_2021,paullada_data_2021,peng_mitigating_2021}, we argue that dataset creators and competition organizers should publish their labeling instructions, as proposed by Maier-Hein et al.~\cite{maier-hein_bias_2020}. 
To advance current practice, we recommend the dataset and labeling instruction creation to be an iterative process properly modeling the underlying distribution of the data space, as described in Freeman et al.~\cite{freeman_iterative_2021}. In earlier stages of the dataset creation process, the focus should be on common occurrences (e.g. common surgical instruments in the case of laparoscopic surgery) to generate strong initial model performance. Throughout the process, special, conflicting or rare occurrences should be added to the dataset to maximize the model performance and reflect the real-world distribution. 

A further recommendation motivated by our study is for annotation requesters to evaluate their annotator options more carefully and select their provider based on suitability to their annotation requirements. While medical personnel alone may be too sparse and costly to satisfy the rising demand for annotated biomedical image data~\cite{freeman_iterative_2021}, oftentimes, medical domain knowledge may only be necessary for the creation of the labeling instructions and not the annotation itself. Thus, crowdsourcing in combination with computer assisted annotation strategies can be a valid and cost-effective approach. \ac{MTurk}, the most commonly used crowdsourcing platform in health research~\cite{crequit_mapping_2018}, follows a do-it-yourself model, where all components from annotator training to annotation tooling are provided by the annotation requester and the crowdworkers are employed on a freelance basis. In contrast, professional annotation companies assign a dedicated contact person that oversees the project, with annotators trained and directly employed by the annotation company. While \ac{MTurk} can quickly scale up with a large number of people, professional annotation companies tend to scale up more slowly. However, professional annotators work on data annotation for a longer proportion of their workday and are usually assigned full-time to an annotation project (cf. Fig.~\ref{fig:work_characteristics}). Regarding location, the international annotation market leads to scenarios where requesters and annotators do not share the same (native) language. This reinforces the need for clear and concise labeling instructions with exemplary pictures as a fundamental requirement for scaling data annotation operations.

Of note, with performance assessment of crowdworkers still conducted by examining their performance on reference standard datasets~\cite{freeman_iterative_2021,lampert_empirical_2016} or measuring the inter-worker agreement~\cite{lendvay_crowdsourcing_2015,nowak_how_2010}, the recommended iterative labeling instruction generation would potentially reduce errors resulting from lacking quality in labeling instructions and should thus result in a more precise assessment of crowdworkers’ actual performances.

A limitation of our study could be seen in the fact that we only included one dataset. Our chosen dataset combines the advantages of high-quality reference annotations, representing the real-world complexity with gradually increasing stages of difficulty and high volume. We chose to only include one sample scenario since running several experiments with the same annotation companies and different datasets poses significant risks of exposing the experimental setting. Annotation companies are typically well aware of the most common datasets and would additionally spot the known experimental structure of gradually increasing labeling instructions within the same layout. Awareness of the experimental setting would in turn lead to results being skewed in favor of high-quality performances, since these companies are inherently motivated to present their work as reliable. A further limitation of our study is that \ac{MTurk} required a different annotation tooling than the annotation tooling used by the professional annotation companies. To mitigate a potential impact of the toolings, all participating annotators had no prior experience with their respective tooling and both toolings included best design practices to enable high quality annotations.

Our study was subject to several design choices. For the performance measurement, we focused on the \ac{DSC} as an overlap-based metric. Utilizing distance-based metrics, such as the Normalized Surface Distance, yielded similar results. For the statistical analysis, we assumed that the non-zero \ac{DSC} scores follow a beta distribution as it is the more natural distribution for this kind of data. We further assumed that the random effects in the two-part \ac{ZIBMM} are normally distributed and correlated. The correlation assumption was reasonable in that, as the probability of severe annotation errors increases, the expected \ac{DSC} score decreases and vice versa.

Even though this paper focuses on labeling instructions associated with biomedical image analysis, we believe that the findings can be translated to other research fields, and to crowdsourced data annotation in general. We expect the impact to go beyond academia because industrial production \ac{ML} projects by nature depend on a predefined level of annotation quality in order to obtain the required algorithm performance level. Consequently, it stands to reason that more effort and monetary resources should especially be invested in developing labeling instructions in industry~\cite{freeman_iterative_2021}.

The present work opens up several future research avenues: First, professional annotation companies usually conduct quality assurance checks by experienced annotators or team leads before providing their created annotations to the annotation requester. Although we obtained significant quality improvements in the obtained annotations by optimizing the labeling instructions, it would be of interest to analyze potential further impact of such quality assurance checks. Additionally, the interaction effects between experienced annotators working on quality assurance checks and instructions with varying levels of information density could be of interest to the scientific community. Second, data pipelines with a long-term focus face the risk of concept drift, where the initially captured distribution of input data changes. How labeling instructions should evolve along ever-changing data and its distribution remains an open question to be tackled. Finally, with an increasing educational shift towards digitalization and data science, involving medical students in medical image annotation as part of their study program could potentially become a new source of crowdsourcing. Future research should examine this promising symbiotic relationship, where medical students obtain hands-on \ac{ML}-related skills highly relevant to their profession while at the same time easing the annotation bottleneck for the scientific community.

In summary, our study is the first to examine the impact of the quality of labeling instructions on annotation work performed both by professional annotators and crowdworkers. We uncovered a significant discrepancy between the demand of professional annotators for better labeling instructions and current common practice. Given the rapidly increasing complexity and diversity of datasets, we envision the establishment and widespread adoption of quality standards for labeling instructions to become imperative in the future.

\section*{Method}
\label{sec:method}

Following the definition of terms used throughout this study, we will describe the selected data, the annotation providers, the labeling instructions, the experimental setup, and the statistical analysis in detail throughout this section.

\subsection*{Definitions}
We use the following terms throughout the paper:

Annotation provider: An entity that provides annotations performed by human workers. They can be categorized into two types of annotation providers: a) crowdsourcing platforms, such as \ac{MTurk}, and b) professional annotation companies (see the following definitions).

Annotation requester: An entity that wants a dataset annotated by an external annotation provider.

Challenge/Competition: We follow the \textit{challenge} definition of the BIAS statement, which defines a challenge as an “open competition on a dedicated scientific problem in the field of biomedical image analysis. A challenge is typically organized by a consortium that issues a dedicated call for participation. A challenge may deal with multiple different tasks for which separate assessment results are provided. For example, a challenge may target the problem of segmentation of human organs in computed tomography (CT) images. It may include several tasks corresponding to the different organs of interest.“~\cite{maier-hein_why_2018}

Challenge/Competition task: The BIAS statement defines a \textit{challenge task} as a “subproblem to be solved in the scope of a challenge for which a dedicated ranking/leaderboard is provided (if any).”~\cite{maier-hein_why_2018}

Labeling instruction: A document or tool that specifies how to correctly label (imaging) data. The different types of labeling instructions are defined below and presented in greater detail in Supp.~\ref{supp:li_type_one}-\ref{supp:li_type_three}.

\ac{MTurk}: A two-sided crowdsourcing marketplace that enables annotation requesters to hire freelance workers, referred to as \ac{MTurk} crowdworkers, to perform discrete tasks on-demand.

\ac{MTurk} crowdworker: A remotely located person performing discrete on-demand tasks on \ac{MTurk} on their own hardware. They are employed on a freelance basis.

Professional annotation company: A company focusing mainly on generating annotations for (imaging) data. Their workers are located in regular office space and mainly employed full-time.

Professional annotator: An on-site located and full-time employed person performing annotations for a professional annotation company in a provided office space with according hardware.

\subsection*{International professional annotator survey}
To obtain a comprehensive understanding of the current issues professional annotators face with respect to labeling instructions and their work characteristics, we developed a 26-item questionnaire (provided in Supp.~\ref{supp:survey_questions}). The survey was distributed among five internationally operating annotation companies in best-cost countries that exclusively employ professional annotators. To increase the statistical validity of the submitted entries (n = 363), we employed a twofold filtering strategy of the entries: a) a control question pair, which consists of the positive and negative formulation of a question, and b) an instructional manipulation check, as recommended by Oppenheimer et al.~\cite{oppenheimer_instructional_2009}. The check asks the participant to answer an open-ended question with a specific set of words that can only be answered by carefully reading the question text. The filtering resulted in 298 remaining entries.

\subsection*{Competition analysis}
Our goal was to capture the current handling of labeling instructions in biomedical image analysis. Thus, we included all \ac{MICCAI} registered competitions\footnote{We retrieved the registered competitions from the \ac{MICCAI} website, where all \ac{MICCAI} registered competitions are published.} which were published until the end of 2021. This resulted in a list of 53 competitions with 96 competition tasks. Two engineers with a proven history in reviewing competitions rated all submitted standardized competition design documents of the individual competition tasks as to whether a labeling instruction was provided by the competition task. In addition, ambiguous cases were marked as such. This resulted in an inter-rater agreement of 90.6\%. Contradictionary ratings were mainly cases where both raters marked the competition task as ambiguous and were solved by an independent third engineer with a proven history in reviewing competitions. 20 competition tasks were excluded as not applicable in the process, as they provided a valid reasoning why labeling instructions cannot be provided. An example is the Medical Out-of-Distribution Analysis Challenge~\cite{zimmerer_mood_2022}, where the publication of the labeling instruction would enable cheating, because it contains information about the placement of out-of-distribution objects in the competition data. The result of the competition tasks analysis is provided in Supp.~\ref{supp:miccai_challenge_analysis_results}.

\subsection*{Dataset selection}
The \ac{HeiCo} segmentation~\cite{maier-hein_heidelberg_2021,ros_comparative_2021} dataset, comprising medical instrument segmentations in laparoscopic video data, served as the basis for this study. Each image was enhanced by 21 meta annotations representing relevant image characteristics or artifacts (e.g. whether overlapping instruments or motion blur are present on the image), where each image characteristic is a binary annotation decision~\cite{ros_how_2021}. The meta annotations were implemented by a trained engineer with extensive annotation experience, who was involved in the original creation process of the dataset. Based on the meta annotations, the images were categorized into nine different image categories, which represented the potential annotation difficulty and served for the subsequent image selection:
\setlist[description]{font=\normalfont\space}
\begin{description}
\item[1) Simple category:] Images do not contain any artifact on the instruments. 
\item[2) Chaos category:] Images contain at least three different artifacts on the instruments. Moreover, images containing a higher number of instruments are preferred.
\item[3) Trocar category:] At least one trocar is present on the image.
\item[4) Intersection category:] At least two medical instruments are intersecting on the image.
\item[5) Motion blur category:] A minimum of one medical instrument with the motion blur artifact is present on the image.
\item[6) Underexposure category:] At least one medical instrument on the image is underexposed.
\item[7) Text overlay category:] Text overlay is present and obstructs the view of the image.
\item[8) Image overlay category:] An image overlay is present and obstructs the view of the image.
\item[9) Random category:] Images are randomly selected from the remaining images in the test set.
\end{description}
We selected 234 unique frames corresponding to the defined categories. Each unique image was annotated four times per labeling instruction and per annotation provider, resulting in 60 annotations per image (generating a total of 14,040 annotated images). 15 unique images from category 1 to category 8 were selected by hand, accounting for roughly half of the images. The only exception was category 8, because there existed only eleven unique images that matched the definition of the category. The other half was selected randomly (category 9).

\subsection*{Annotation providers}
The study was conducted based on five annotation providers, consisting of four professional annotation providers and the crowdsourcing platform \ac{MTurk}. Each annotator participated exclusively with one of the three labeling instructions. We selected high quality representatives for the professional annotation companies and the crowdsourcing platform. The professional annotation companies operate internationally and their annotators are located in best-cost countries. The selected companies had a proven track record in large-scale industry annotation projects. Participating crowdworkers on \ac{MTurk} had to fulfill the following quality requirements: a) 98\% accepted \ac{HITs} and b) a minimum of 5,000 accepted HITs. Our quality requirements thus far surpassed the quality requirements of researchers working with \ac{MTurk}, which normally require 95\% accepted HITs with a minimum of 100 accepted \ac{HITs}~\cite{kennedy_shape_2020,heim_large-scale_2018,bragg_sprout_2018,cheplygina_early_2016}. To capture a representative sample of the population on \ac{MTurk}, we spread our \ac{HITs} across 40 days and all times of day. We followed Litman et al. to ensure a fair worker compensation~\cite{litman_relationship_2015}.

\subsection*{Labeling instructions}
As a labeling instruction specifies how to correctly label the (imaging) data, we defined a set of design rules that are shared across all labeling instructions:
\begin{enumerate}[label=(\alph*)]
\item The information of a labeling instruction is provided in a slide layout for better human information processing.
\item Each slide represents a chunk, an encapsulated unit of information. Chunking reduces the demand on the working memory.
\item Related information chunks are positioned near each other.
\item A consistent layout with defined fonts, symbols and colors is applied.
\end{enumerate}

\textit{Minimal text labeling} instructions:
The \textit{minimal text labeling} instructions consist of a limited textual description, including positive examples and the most common annotation occurrences (see Supp.~\ref{supp:li_type_one}). This represents a situation where only little effort was put into creating the labeling instruction. For example, text overlay references the uncommon occurrence of text that is visible in the image. Because it is an uncommon occurrence, it is not mentioned in the \textit{minimal text labeling} instruction. As a baseline, the \textit{minimal text labeling} instructions consist of seven slides with 168 words. 

\textit{Extended text labeling} instructions:
The \textit{extended text labeling} instructions extend the \textit{minimal text labeling} instructions with a comprehensive text description, which is supported by both positive examples and counterexamples in text form. Furthermore, both common and uncommon cases are included, see Supp.~\ref{supp:li_type_two}. In this labeling instruction for example, the uncommon occurrence of text overlay is described in detail. This resulted in ten slides with 446 words.

\textit{Extended text including pictures labeling} instructions:
The \textit{extended text including pictures labeling} instructions complement the \textit{extended text labeling} instructions with pictures, see Supp.~\ref{supp:li_type_three}. The pictures include textual descriptions, symbols, markings and the usage of color to convey the information on the slides. In addition, rare annotation occurrences are included as well. This represents a situation where extensive (domain) knowledge about the labeling process is present and documented in detail in the labeling instruction. Hence, in our example, the uncommon occurrence of text overlay is described in detail with text and pictures. These labeling instructions consist of 16 slides with 961 words.

\subsection*{Experimental setup for quantifying the impact of labeling instructions}
Each of the five labeling providers annotated the same images subsequently with each labeling instruction, using separate annotators. We started with the \textit{minimal text labeling} instructions, followed by the \textit{extended text labeling} instructions and provided the \textit{extended text including pictures labeling} instructions last, to prevent information leakage. As an additional security measure, we added a minimum break of ten days between two labeling instructions. No individual questions from the annotators regarding the labeling instruction content were answered in order to prevent a potential information advantage for an annotation provider that could impact the statistical analysis. Each annotator was only allowed to participate for a single labeling instruction. All participating annotators had no prior experience with their respective annotation tooling. Furthermore, no worker selection tasks were used for the annotation providers. Each professional annotator annotated a total of 72 images. To properly simulate the parallelization of crowdsourcing, each \ac{MTurk} crowdworker annotated four images. After the submission of their annotations, each \ac{MTurk} crowdworker could submit a short optional survey about their work characteristics.

\subsection*{Statistical analysis}
To quantify the impact of labeling instructions and the two annotator types, the following statistical methods were used: 

To ensure compatibility with prior work on the data, we utilized the same metrics as suggested by Roß et al.~\cite{ros_comparative_2021}, in which the dataset was originally introduced as part of the \ac{MICCAI} Robust Medical Instrument Segmentation Challenge 2019. We analyzed the annotation results based on the \ac{DSC} scores with a two-part zero-inflated beta mixed model~\cite{chen_two-part_2016}: a) the first part included a logistic mixed model analyzing the probability of severe annotation errors (at least one instrument with \ac{DSC} = 0 in one frame) and b) the second part consisted of a beta mixed model analyzing the non-zero \ac{DSC} values of an image when valid annotations occurred. The image variable and the annotation worker variable were modeled as random effects while the type of labeling instructions, annotator type, image category and access to context video were modeled as fixed effects. The model was implemented in the brms package in R~\cite{r_language_reference}, where vague Gaussian priors centered on 0 were used for the fixed effects and Half-Cauchy priors were assigned for the standard deviation of the random effects. 4,000 Markov Chain Monte Carlo samples were generated across 4 chains. The obtained estimates of the covariates are on the log-odds scale and were exponentiated to obtain the odds ratio for each covariate.
Software: R version 4.0.2 (package brms version 2.16.0). 

\newpage
\appendix
\section*{Data availability}
\label{sec:data_availability}
Five datasets were utilized during the current study: 
DS1: Captured biomedical competition design documents from publicly available sources (2020-2021).
DS2: Reference annotations for the \ac{HeiCo}~\cite{maier-hein_heidelberg_2021,ros_comparative_2021} dataset.
DS3: Captured annotations from professional annotators and \ac{MTurk} crowdworkers. 
DS4: Individual professional annotator responses to the survey "Labeling Instructions Survey".
DS5: Individual MTurk crowdworker responses to optional work characteristics survey. These questions are a subset of the DS4 questions.
For DS1, the individual challenge design documents are freely available at MICCAI~\cite{noauthor_miccai_nodate}.
A reporting summary for the evaluation is available as Supp.~\ref{supp:miccai_challenge_analysis_results}.
DS2 is freely available from Synapse~\cite{noauthor_robust_nodate}.
DS3, DS4 and DS5 are available from the corresponding author L.M.-H. upon reasonable request.

\appendix
\section*{Code availability}
\label{sec:code_availability}
The code for the statistical analyses, and generating the metric scores between the reference annotations and the captured annotations is available from the corresponding author L.M.-H. upon reasonable request.

\section*{Acknowledgements}
understand.ai, Karlsruhe provided the annotations for this work and funded a part of this work, namely Marc Mengler and Dr. Simon Funke. A part of this work was funded by Helmholtz Imaging (HI), a platform of the Helmholtz Incubator on Information and Data Science.
We would like to thank Matthias Eisenmann for his continuous feedback. Furthermore, we want to thank Konstantin D. Pandl, Ali Sunyaev and the Karlsruhe Institute of Technology (KIT), where a part of this research was conducted.


\section*{Ethics declarations}
\subsection*{Competing interests}
Tim Rädsch was an employee of the company understand.ai, which sponsored the creation of the annotations. 
After his research, Tobias Ross was employed by Quality Match GmbH.

\bibliographystyle{ACM-Reference-Format}
\bibliography{sample-base}
\newpage

\section*{Acronyms}
\begin{acronym}[]
\acro{CI}{Credible Interval}
\acro{CT}{computed tomography}
\acro{DL}{Deep Learning}
\acro{DSC}{Dice Similarity Coefficient}
\acro{HeiCo}{Heidelberg Colorectal}
\acro{HITs}{Human Intelligence Tasks}
\acro{IQR}{Interquartile Range}
\acro{ML}{Machine Learning}
\acro{MICCAI}{Medical Image Computing and Computer Assisted Intervention Society}
\acro{MTurk}{Amazon Mechanical Turk}
\acro{ZIBMM}{zero-inflated beta mixed model}
\end{acronym}

\newpage
\appendix
\section*{Supplementary}
\label{sec:supplementary}
\addcontentsline{toc}{section}{Supplementary}
\renewcommand{\thesubsection}{\Alph{subsection}}

\subsection{PRISMA statement}
\label{supp:prisma_statement}
\begin{figure}[H]
    \centering
    \includegraphics[width=\linewidth]{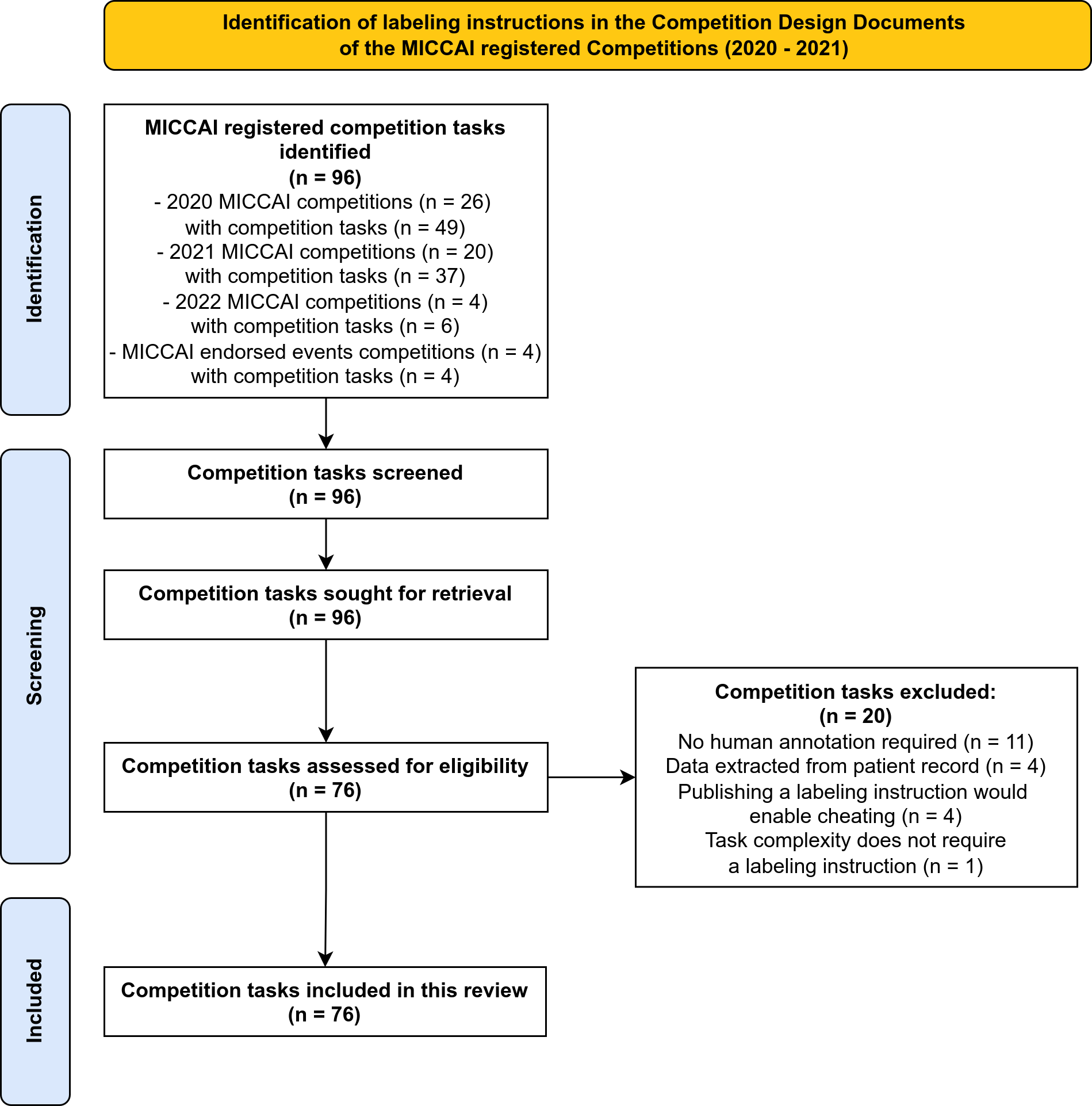}
    \caption{\textbf{Structured identification of labeling instructions in the Challenge Design Documents of the \acl{MICCAI} (\acs{MICCAI}) registered competitions (2020 - 2021). }After the identification of all relevant competition tasks, each individual competition task was screened and excluded, if the organizers provided a valid reasoning why their labeling instructions cannot be provided.}
    \label{fig:prisma_statement}
\end{figure}
\newpage

\subsection{DSC scores for the category with the least and the most present image characteristics on the instrument level}
\label{supp:dsc_simple_and_chaos_category}
\begin{figure}[H]
    \centering
    \includegraphics[width=\linewidth]{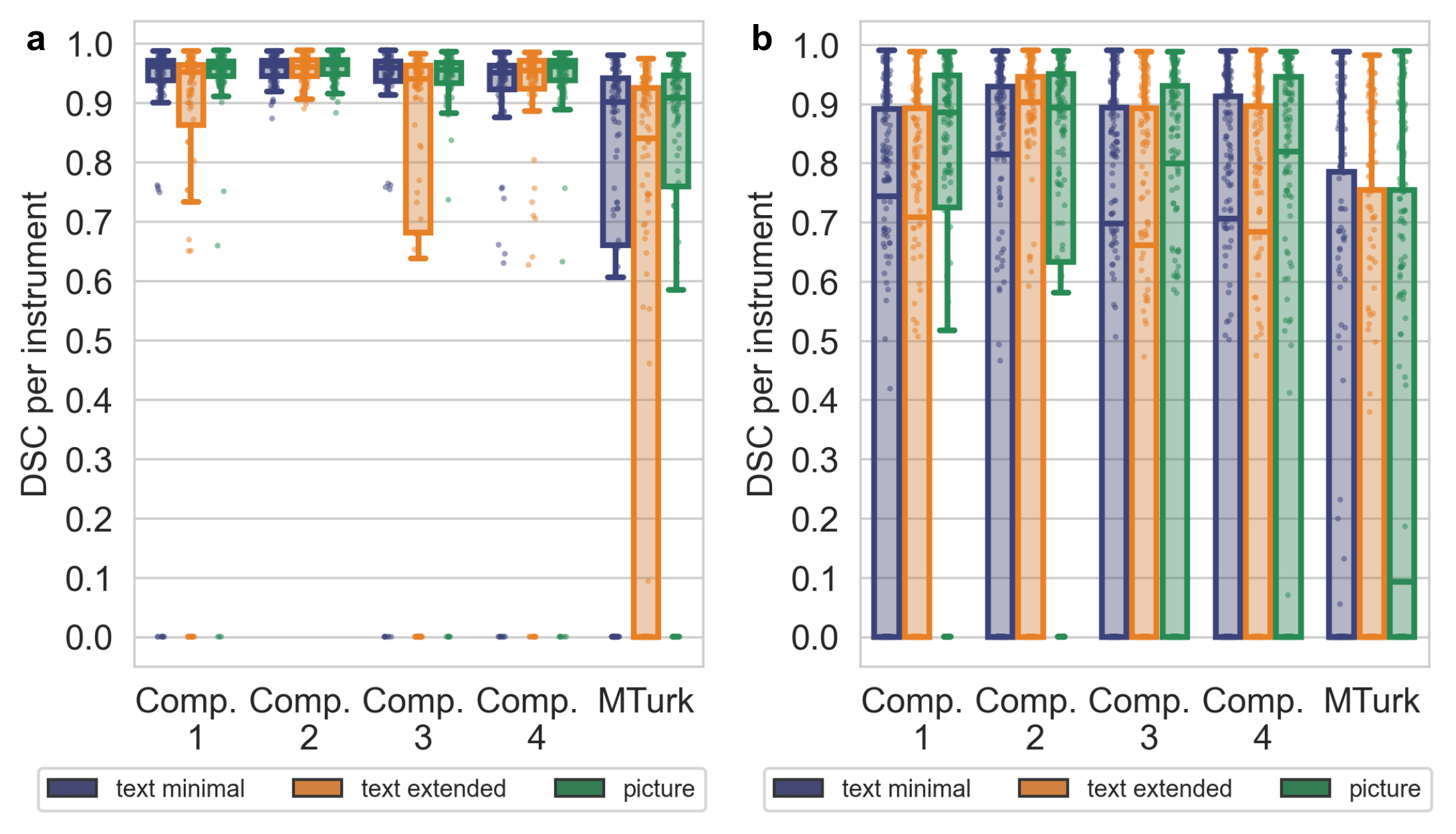}
    \caption{\textbf{\ac{DSC} scores for the category with the least and the most image characteristics present on the instrument level.} For each annotation provider and type of labeling instruction, the \acl{DSC} (\acs{DSC}) scores per instrument are aggregated. \textbf{a} displays the category with images that do not contain any artifact on the instruments. \textbf{b} displays the category with images that contain at least three different artifacts on the instruments. The \ac{DSC} scores are displayed as a dots- and boxplot (the band indicates the median, the box indicates the first and third quartiles and the whiskers indicate $ \pm 1.5 \times$ interquartile range). A combined total of 3771 instrument annotations were observed for the category with the least and the most present image characteristics.}
    \label{fig:dsc_simple_and_chaos}
\end{figure}
\newpage

\subsection{Aggregated MICCAI competition analysis results}
\label{supp:miccai_challenge_analysis_results}
\begin{table}[!htp]\centering
\caption{\textbf{Aggregated results of the \acl{MICCAI} (\acs{MICCAI}) competition analysis. }For each competition, the individual competition task submission documents were evaluated.}
\label{tab: }
\small
\begin{tabular}{p{4cm}rrr}\toprule
\multicolumn{4}{c}{\textbf{Analysis results of MICCAI registered competitions}} \\\midrule
\textbf{Date of evaluation} & Nov 10 2021 & & \\
\textbf{Total number of competitions*} &53 & & \\
\textbf{Total number of competition tasks} & 96 & & \\
 & & & \\
\textbf{Competition tasks which..} &\textbf{absolute} &\textbf{relative} & \\
\textbf{do not need to publish LIs** or have valid justification} &20 &20.83\% & \\
\textbf{Could/should publish} &76 &79.17\% & \\
& & & \\
\textbf{In scope competition tasks (76) which ...} &\textbf{absolute} &\textbf{relative} & \\
\textbf{publish LIs} &18 &23.68\% & \\
\textbf{do not publish / have LIs} & 58 & 76.32\% & \\
\textbf{Inter-rater agreement:} &90.63\% & & \\
& & & \\
\textbf{} & & & \\
** LI = Labeling Instruction & & & \\
\multicolumn{4}{l}{* All information based on the Challenge design docs, mainly question 23 b/a, of the}\\
\multicolumn{4}{l}{MICCAI registered competitions (\href{https://miccai.org/index.php/special-interest-groups/challenges/miccai-registered-challenges/}{Link} | accessed: 2021-11-10) between 2020 and 2021.}\\
\bottomrule
\end{tabular}
\end{table}
\newpage



\section*{Supplementary material}

## D  Survey: Current status of labeling instructions

# Labeling Instructions Survey

## Section 1: Introduction

This survey aims to collect information about people who perform annotations. Labeling instructions provide the information on "what and how to annotate".

It will take around 5 minutes to finish the survey.
The collected information will help us improve labeling instructions in the future.
The collected information is anonymous and will not be shared with your employer.

This survey consists of two parts:
- General questions
- Annotation-related questions

contact information: xxxx

I hereby confirm that my responses will be processed in an anonymous fashion in a study on labeling instructions.
- Yes

## Section 2: General questions

What device are you using for the labeling?

Question type: Multiple choice
- Desktop PC
- Laptop

How old are you?

Question type: Drop-down
- 15-20 years
- 21-30 years
- 31-40 years
- 41-50 years
- 50-60 years
- 61-70 years
- Other

Please rate your English skills.

Question type: Drop-down
- Beginner
- Elementary
- Intermediate

- Advanced
- Proficient
- Native speaker

What is your native language?

Question type: Open text
> ➢ YOUR ANSWER

How long have you been using a computer for?

Question type: Drop-down
- 0-5 years
- 6-10 years
- 11-15 years
- 16-20 years
- More than 20 years

What is your highest earned degree?

Question type: Drop-down
- Basic education
- Higher education (similar to high school)
- Bachelor - University or similar
- Master - University or similar
- Doctor - University or similar
- None of the above

## Section 3: Annotation-related questions

How often do you label per week?

Question type: Drop-down
- 1 day per week
- 2 days per week
- 3 days per week
- 4 days per week
- 5 days per week
- More than 5 days per week

Which of the mentioned options cause problems in the daily annotation work of your colleagues? You can choose multiple answers.

Question type: Checkboxes
- Missing training
- Unclear labeling instructions / Unclear labeling guide
- Tooling issues / Tooling restrictions (e.g. not able to save annotations)
- Poor data (e.g. image quality or sensor fusion)
- Concentration issues (due to monotonous work)
- Other…

Is labeling your main source of income?

Question type: Multiple choice
- Yes
- No

How many hours do you spend per week on labeling images? Please note that all activities regarding labeling (e.g. training) are counted here.

Question type: Drop-down
- 1-2 hours
- 3-5 hours
- 6-10 hours
- 11-15 hours
- 16-20 hours
- 21-30 hours
- 31-40 hours
- more than 40 hours

How long have you been labeling for?

Question type: Drop-down
- Less than 1 year
- Between 1 and 1.5 years
- Between 1.6 and 2 years
- Between 2.1 and 2.5 years
- Between 2.6 and 3 years
- Between 3.1 and 4 years
- Between 4 and 5 years
- Between 6 and 7 years
- Between 8 and 9 years
- More than 10 years

Please rate the following statements. If you do not understand the statement, you can select "Do not understand statement" for certain statements.

I prefer short labeling instructions rather than long labeling instructions.

Question type: Multiple choice (Likert Scale)
- Strongly disagree
- Disagree
- Neither agree nor disagree
- Agree
- Strongly agree

Labeling instructions are not important for annotation projects.

Question type: Multiple choice (Likert Scale)
- Strongly disagree
- Disagree
- Neither agree nor disagree

- Agree
- Strongly agree

After reading more detailed labeling instructions my colleagues are able to generate annotations faster than with less detailed labeling instructions.

Question type: Multiple choice (Likert Scale)
- Strongly disagree
- Disagree
- Neither agree nor disagree
- Agree
- Strongly agree
- Do not understand statement

After reading more detailed labeling instructions my colleagues are able to generate annotations with higher quality than with less detailed labeling instructions.

Question type: Multiple choice (Likert Scale)
- Strongly disagree
- Disagree
- Neither agree nor disagree
- Agree
- Strongly agree
- Do not understand statement

The creator of the labeling instructions should invest more time and resources in the generation of the labeling instructions.

Question type: Multiple choice (Likert Scale)
- Strongly disagree
- Disagree
- Neither agree nor disagree
- Agree
- Strongly agree
- Do not understand statement

Please rate the following statements. The scale options are defined as:
Never: 0%
Rarely: Between 1% and 25%
Sometimes: Between 26% and 50%
Often: Between 51% and 75%
Always: Between 76% and 100%

Unclear labeling instructions lead to questions of my colleagues about the labeling instructions.

Question type: Multiple choice (Likert Scale)
- Never
- Rarely
- Sometimes

- Often
- Always

In the past, my colleagues made annotation mistakes due to unclear labeling instructions.

Question type: Multiple choice (Likert Scale)
- Never
- Rarely
- Sometimes
- Often
- Always

Extensive text descriptions and written examples in the labeling instructions improve the understanding of the labeling instructions for my colleagues.

Question type: Multiple choice (Likert Scale)
- Never
- Rarely
- Sometimes
- Often
- Always

In the past, my colleagues made annotation mistakes due to incomplete labeling instructions.

Question type: Multiple choice (Likert Scale)
- Never
- Rarely
- Sometimes
- Often
- Always

Pictures in the labeling instructions improve the understanding of the labeling instructions for my colleagues.

Question type: Multiple choice (Likert Scale)
- Never
- Rarely
- Sometimes
- Often
- Always

In the past, I made annotation mistakes due to unclear labeling instructions.

Question type: Multiple choice (Likert Scale)
- Never
- Rarely
- Sometimes
- Often
- Always

29Please rate the following statements. If you do not understand the statement, you can select "Do not understand statement" for certain statements. The scale options are defined as:
Never: 0%
Rarely: Between 1% and 25%
Sometimes: Between 26% and 50%
Often: Between 51% and 75%
Always: Between 76% and 100%

The labeling instructions my colleagues receive for annotation projects are usually of good quality and enable them to generate high quality annotations.

Question type: Multiple choice (Likert Scale)
- Never
- Rarely
- Sometimes
- Often
- Always
- Do not understand statement

Unclear labeling instructions delay the completion of annotation projects.

Question type: Multiple choice (Likert Scale)
- Never
- Rarely
- Sometimes
- Often
- Always
- Do not understand statement

Unclear labeling instructions cause rework of the annotations for my colleagues.

Question type: Multiple choice (Likert Scale)
- Never
- Rarely
- Sometimes
- Often
- Always
- Do not understand statement

Is there anything you would like to add or mention?

Question type: Open text
➢ YOUR ANSWER

Thank you for participating in the survey and helping to improve labeling instructions.

# E  Minimal text labeling instruction

## Overview

- Terminology ............................................................. 2
- Goal........................................................................ 3
- Occlusion................................................................ 4
- Medical instruments............................................... 5
- Medical instruments - Holes.................................. 6
- Medical instruments - Transparency..................... 7

1

## Terminology

**Definition "Matter":**
- Anything that has mass, takes up space and can be clearly identified.
- Examples: tissue, surgical tools, blood

2

## Goal - Detect and segment all relevant medical instruments

**Only segment medical instruments**
- Each medical instrument is their own instance.
- Each instance has their own colour.
- The interior of a segmentation represents a medical instrument.
- Everything outside the medical instrument segmentations is regarded as other matter.

3

## Occlusion

Each pixel may correspond to exactly one instance.
The solid/liquid matter that occurs first along the line of sight of the endoscope determines the label.

4

## Medical instruments

**Definition "Medical instrument":**
- An elongated rigid object placed inside the patient and manipulated from outside the patient.
- Examples: grasper, scalpel, (transparent) trocar, clip applicator, hooks, stapling device, suction

5

## Medical instruments - Holes

Several medical instruments feature holes.
They are regarded as part of the instrument.

6

# Medical instruments - Transparency

Medical instruments may be transparent.
The occlusion rule holds in this case as well.

7

## F Extended text labeling instruction

# Overview

- Terminology……………………………………………….. 2
- Goal………………………………………………………….. 3
- Occlusion…………………………………………………… 4
- Medical instruments……………………………………. 5
- Medical instruments - Holes………………………… 6
- Medical instruments - Holes exception………… 7
- Medical instruments - Transparency……………. 8
- Text overlay………………..……………………………… 9
- Image overlay……………..……………..………………. 10

1

# Terminology

**Definition "Matter":**
- Anything that has mass, takes up space and can be clearly identified.
- Examples: tissue, surgical tools, blood
- Counterexamples: reflections, digital overlays, movement artifacts, smoke

2

## Goal - Detect and segment all relevant medical instruments

**Only segment medical instruments**
- Each medical instrument is their own instance.
- If two parts of the same instrument are visible, they belong to the same instrument instance.
- Each instance has their own colour.
- The interior of a segmentation represents a medical instrument.
- Everything outside the medical instrument segmentations is regarded as other matter.
- Be as precise as possible. Every pixel counts in surgery.

3

## Occlusion

Each pixel may correspond to exactly one instance.
The solid/liquid matter that occurs first along the line of sight of the endoscope determines the label.

This may result in multiple segmentations for a single instrument that is occluded by another instrument, blood or tissue, for example.

4

## Medical instruments

**Definition "Medical instrument":**
- An elongated rigid object placed inside the patient and manipulated from outside the patient.
- Examples: grasper, scalpel, (transparent) trocar, clip applicator, hooks, stapling device, suction

- Counterexamples: non-rigid tubes, bandage, compress, needle (not directly manipulated from outside but manipulated with an instrument), coagulation sponges, metal clips

5

## Medical instruments - Holes

Several medical instruments feature holes.
They are regarded as part of the instrument.

A hole is made up of pixels that do not show

parts of the instrument but are either

    type a) completely surrounded by pixels of the same instrument or

    type b) completely surrounded by pixels of one instrument and the margin of the instrument would close the hole outside the image.

6

## Medical instruments - Holes exception

The sole exception are trocars when the camera is placed inside them.

Trocars are transparent medical instruments
- They are placed through the abdomen.
- They function as a portal for the other medical instruments.
- When the camera is inside a trocar, a trocar can take up a large part of the image.

Trocars are exempt from the hole rule
- Reason: Otherwise, the holes would make up a large part of the image.

7

## Medical instruments - Transparency

Medical instruments may be transparent.
The occlusion rule holds in this case as well.

8

## Text overlay

Text overlay is text that is visible in the image.
The background of the text is the regular image.

Text overlay shall be ignored.

9

## Image overlay

Image overlay is often uniform coloured shapes
that overlay with the image.
Text with an added uniform color background
is considered as image overlay.

Image overlay is treated like "other matter" and should not be part of the segmentation.

10

## G  Extended text including pictures labeling instruction

# Overview

1

# Terminology

**Definition "Matter":**
- Anything that has mass, takes up space and can be clearly identified.
- Examples: tissue, surgical tools, blood
- Counterexamples: reflections, digital overlays, movement artifacts, smoke

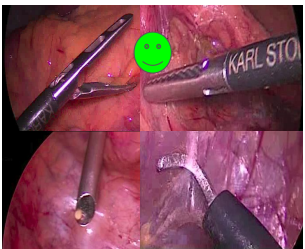
Examples of matter:
Surgical tools.

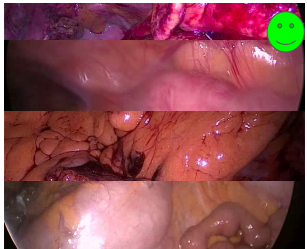
Examples of matter:
Tissue.

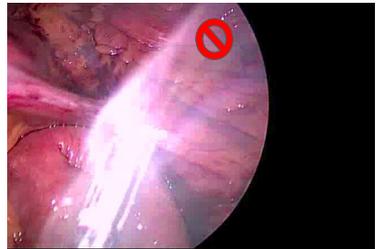
Counterexample:
Smoke.

2

## Goal - Detect and segment all relevant medical instruments

**Only segment medical instruments**
- Each medical instrument is their own instance.
- If two parts of the same instrument are visible, they belong to the same instrument instance.
- Each instance has their own colour.

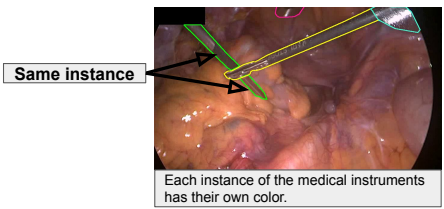
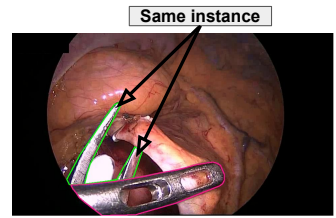

Each instance of the medical instruments has their own color.

3

## Goal - Detect and segment all relevant medical instruments

**Only segment medical instruments**
- The interior of a segmentation represents a medical instrument.
- Everything outside the medical instrument segmentations is regarded as other matter.
- Be as precise as possible. Every pixel counts in surgery.

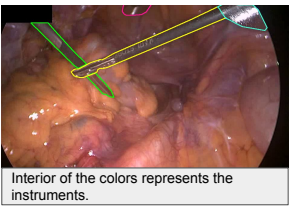
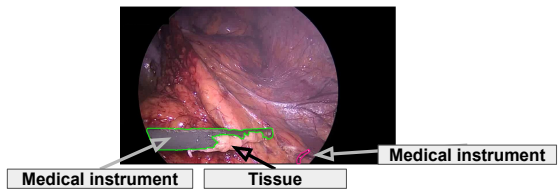

Interior of the colors represents the instruments.

4

# Occlusion

Each pixel may correspond to exactly one instance.

- The solid/liquid matter that occurs first along the line of sight of the endoscope determines the label.

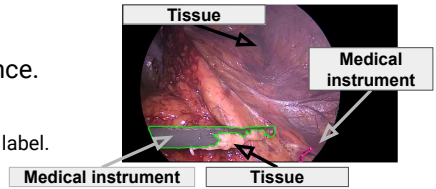

This may result in multiple segmentations for a single instrument that is occluded by another instrument, blood or tissue, for example.

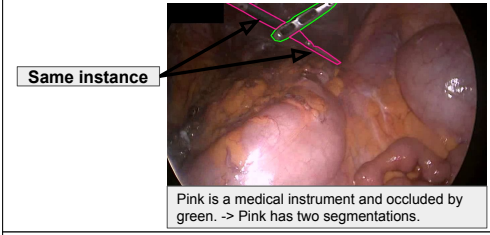

Pink is a medical instrument and occluded by green. -> Pink has two segmentations.

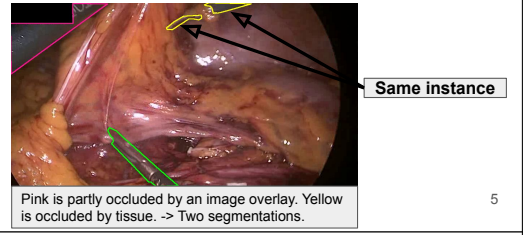

Pink is partly occluded by an image overlay. Yellow is occluded by tissue. -> Two segmentations.

5

# Medical instruments

**Definition "Medical instrument":**
- An elongated rigid object placed into the patient and manipulated from outside the patient.
- Examples: grasper, scalpel, (transparent) trocar, clip applicator, hooks, stapling device, suction

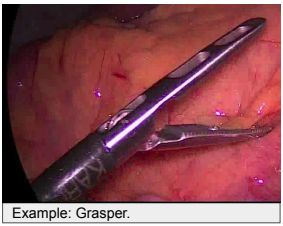
Example: Grasper.

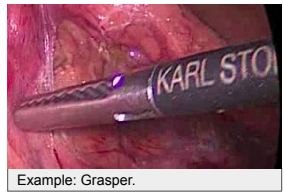
Example: Grasper.

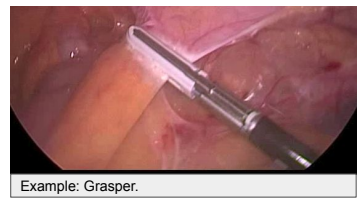
Example: Grasper.

6

## Medical instruments

**Definition "Medical instrument":**
- An elongated rigid object placed into the patient and manipulated from outside the patient.
- Examples: grasper, scalpel, (transparent) trocar, clip applicator, hooks, stapling device, suction

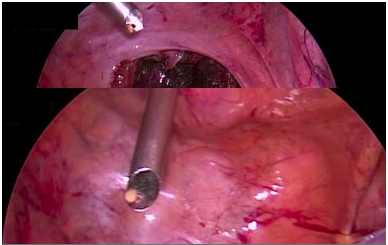
Example: Suction.

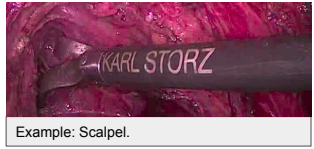
Example: Scalpel.

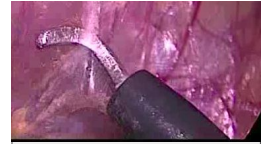
Example: Spreader.

7

## Medical instruments

**Definition "Medical instrument":**
- An elongated rigid object placed into the patient and manipulated from outside the patient.
- Examples: grasper, scalpel, (transparent) trocar, clip applicator, hooks, stapling device, suction

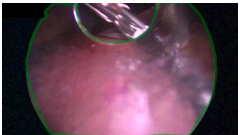
Example: Inside trocar (green). Another instrument is visible through the hole.

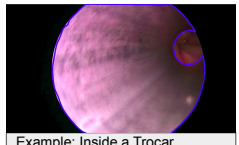
Example: Inside a Trocar.

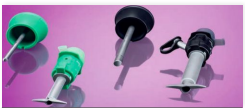
Example: Different trocars.

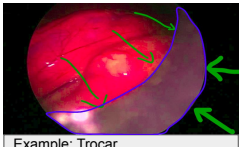
Example: Trocar.

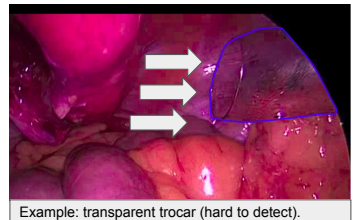
Example: transparent trocar (hard to detect).

8

# Medical instruments

**Definition "Medical instrument":**

- Counterexamples: non-rigid tubes, bandage, compress, needle (not directly manipulated from outside but manipulated with an instrument), coagulation sponges, metal clips

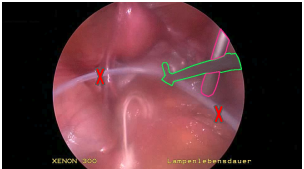 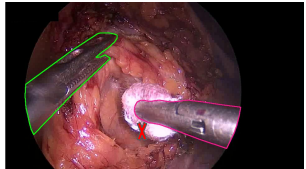 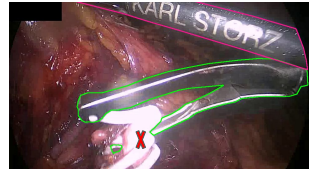

Green and pink are medical instruments.
The red cross marks a drain-plastic
-> no medical instrument.

Green and pink are medical instruments.
The red cross marks a bandage
-> no medical instrument.

Green and pink are medical instruments.
The red cross marks a clamp
-> no medical instrument.

9

# Medical instruments

**Definition "Medical instrument":**

- Counterexamples: non-rigid tubes, bandage, compress, needle (not directly manipulated from outside but manipulated with an instrument), coagulation sponges, metal clips

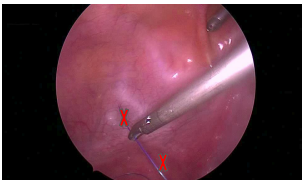 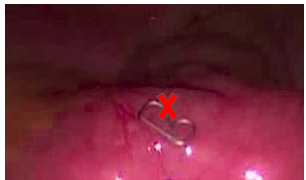 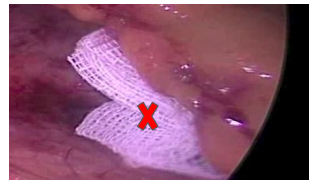

2 instruments are visible.
The red cross marks a needle
-> no medical instrument.

0 instruments are visible.
The red cross marks a metal clip
-> no medical instrument.

0 instruments are visible.
The red cross marks a bandage
-> no medical instrument.

10

## Medical instruments - Holes

Several medical instruments feature holes. They are regarded as part of the instrument.

A hole is made up of pixels that do not show parts of the instrument but are either

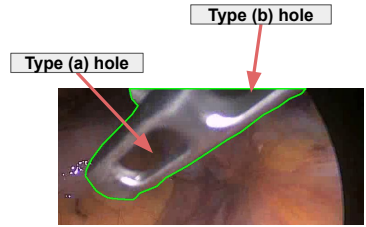

type (a) completely surrounded by pixels of the same instrument or

type (b) are completely surrounded by pixels of one instrument and the margin of the instrument would close the hole outside the image.

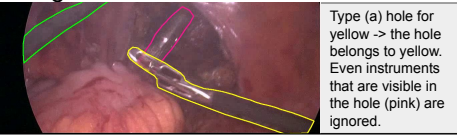

Type (a) hole for yellow -> the hole belongs to yellow. Even instruments that are visible in the hole (pink) are ignored.

11

## Medical instruments - Holes exception

The sole exception are trocars when the camera is placed inside of them.

Trocars are transparent medical instruments.
- They are placed through the abdomen.
- To function as a portal for the other medical instruments.

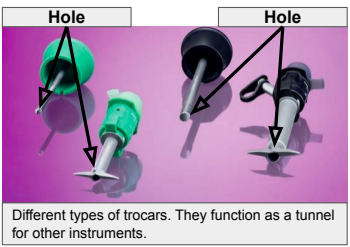
Different types of trocars. They function as a tunnel for other instruments.

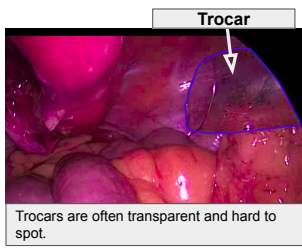
Trocars are often transparent and hard to spot.

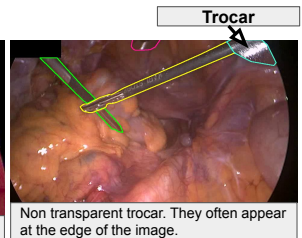
Non transparent trocar. They often appear at the edge of the image.

12

## Medical instruments - Holes exception

The sole exception are trocars when the camera is placed inside of them.

When the camera is inside a trocar, a trocar can take up a large part of the image

Trocars are exempt from the hole rule.
- Otherwise, the holes would contain a large part of the image.

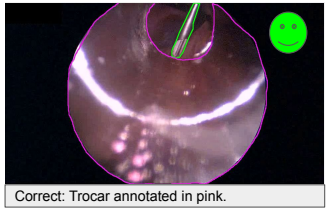
Correct: Trocar annotated in pink.

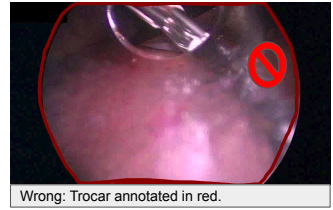
Wrong: Trocar annotated in red.

13

## Medical instruments - Transparency

Medical instruments may be transparent.

The occlusion rule holds in this case as well.

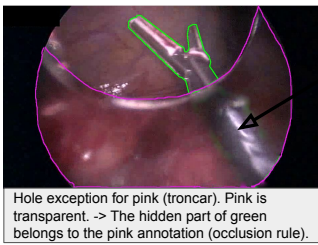
Hole exception for pink (troncar). Pink is transparent. -> The hidden part of green belongs to the pink annotation (occlusion rule).

**Pixels only belong to one instrument.** Here they belong to the transparent instrument, because the transparent instrument is the first object in line of sight (occlusion rule).

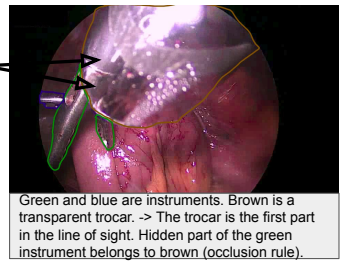
Green and blue are instruments. Brown is a transparent trocar. -> The trocar is the first part in the line of sight. Hidden part of the green instrument belongs to brown (occlusion rule).

14

## Text overlay

Text overlay is text that is visible in the image.
The background of the text is the regular image.

Text overlay shall be ignored.

Examples:

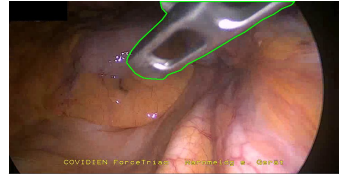

Text overlay (in yellow color).

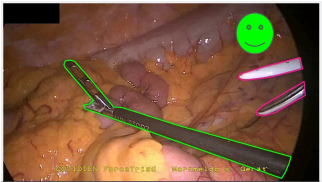

Correct: The text overlay is ignored.

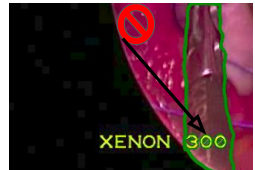

Wrong: The text overlay is treated with a cutout.

15

## Image overlay

Image overlay is often uniform coloured shapes that overlay on the image.
Text with a added uniform color background is considered as image overlay.

Image overlays are treated like "other matter".

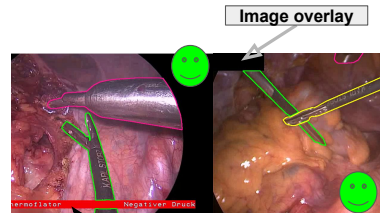

**Image overlay**

**Image overlay**

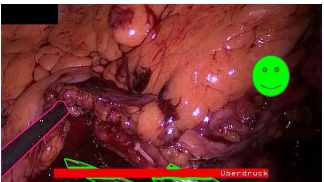

Correct: The image overlay is not part of the instrument.

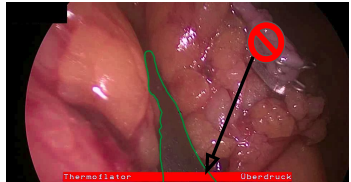

Wrong: The image overlay here is part of the instrument.

16


\newpage

\end{document}